\documentclass{llncs}
\usepackage[utf8]{inputenc}
\usepackage{algorithm}

\usepackage{algpseudocode}
\usepackage{amsmath}
\usepackage{amssymb}
\usepackage{bm}
\usepackage{color}
\usepackage{subfig}
\usepackage{todonotes}
\usepackage{graphicx}
\usepackage{multirow}
\usepackage{breqn}
\usepackage{booktabs}
\usepackage{makecell}
\usepackage{soul}

\title{Reinforced Auto-Zoom Net: Towards Accurate and Fast Breast Cancer Segmentation in Whole-slide Images}
\author{Nanqing Dong
\inst{1,2} 
\and Michael Kampffmeyer
\inst{3}
\and Xiaodan Liang
\inst{4}
\and Zeya Wang
\inst{1}
\and
\\Wei Dai
\inst{1}
\and Eric Xing
\inst{1}
}
\institute{
Petuum, Inc., Pittsburgh, PA 15217, USA \\
\and
Cornell University, Ithaca, NY 14850, USA \\ 
\and
UiT The Arctic University of Norway, 9019 Tromsø, Norway \\
\and
Carnegie Mellon University, Pittsburgh, PA 15213, USA \\
}

\begin{document}
\pagestyle{headings}
\maketitle
\setcounter{footnote}{0}
\begin{abstract}
Convolutional neural networks have led to significant breakthroughs in the domain of medical image analysis. However, the task of breast cancer segmentation in whole-slide images (WSIs) is still underexplored. WSIs are large histopathological images with extremely high resolution. Constrained by the hardware and field of view, using high-magnification patches can slow down the inference process and using low-magnification patches can cause the loss of information. In this paper, we aim to achieve two seemingly conflicting goals for breast cancer segmentation: accurate and fast prediction. We propose a simple yet efficient framework Reinforced Auto-Zoom Net (RAZN) to tackle this task. Motivated by the zoom-in operation of a pathologist using a digital microscope, RAZN learns a policy network to decide whether zooming is required in a given region of interest. Because the zoom-in action is selective, RAZN is robust to unbalanced and noisy ground truth labels and can efficiently reduce overfitting. We evaluate our method on a public breast cancer dataset. RAZN outperforms both single-scale and multi-scale baseline approaches, achieving better accuracy at low inference cost.

\end{abstract}

\section{Introduction}
\begin{figure}[t]
\begin{center}
\captionsetup[subfigure]{labelformat=empty}
\subfloat[(a) normal]{\includegraphics[width=0.22\linewidth,height=0.22\linewidth]{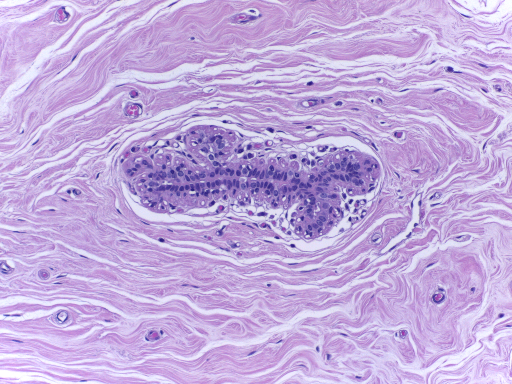}}\vspace{-0.05cm}
\subfloat[(b) benign]{\includegraphics[width=0.22\linewidth,height=0.22\linewidth]{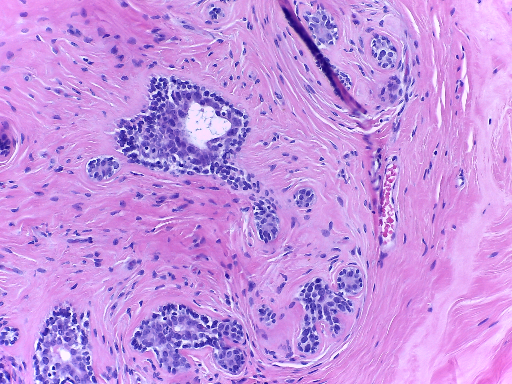}}\vspace{-0.05cm}
\subfloat[(c) in situ]{\includegraphics[width=0.22\linewidth,height=0.22\linewidth]{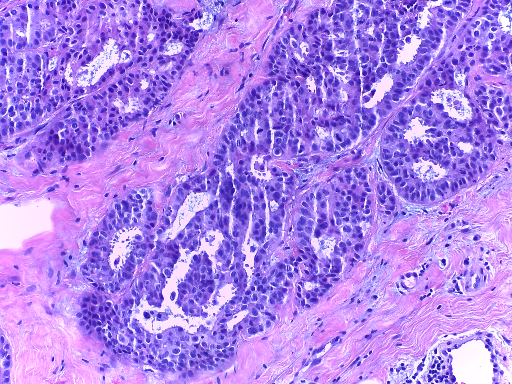}}\vspace{-0.05cm}
\subfloat[(d) invasive]{\includegraphics[width=0.22\linewidth,height=0.22\linewidth]{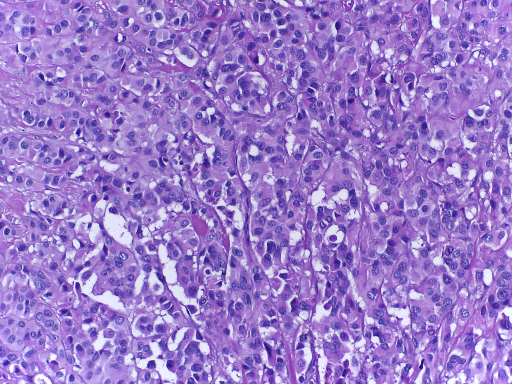}}\vspace{-0.05cm}
\end{center}
   \caption{Examples of different types of tissue. The microscopy images (patches of WSIs at $200\times$ magnification) are labeled according to the predominant tissue type in each image.}
\label{fig:cancer}
\end{figure}

Breast cancer is one of the most common causes of mortality in the female population in the world~\cite{acs2017breast}. It accounts for around $25\%$ of all the cancers diagnosed in women~\cite{alshanbari2015breast}. For traditional diagnostic tools like mammography, even experienced radiologists can miss $10-30\%$ of breast cancers during routine screenings~\cite{cheng2003computer}. With the advent of digital imaging, whole-slide imaging has gained attention from the clinicians and pathologists because of its reliability. Whole-slide images (WSIs) have been permitted for diagnostic use in the USA~\cite{fda}. They are the high-resolution scans of conventional glass slides with Hematoxylin and Eosin (H\&E) stained tissue. There are four types of tissue in breast biopsy: \textit{normal}, \textit{benign}, \textit{in situ carcinoma}, and \textit{invasive carcinoma}. Fig.~\ref{fig:cancer} shows examples of the four types of breast tissue. In clinical testing, the pathologists diagnose breast cancer based on 1) the percentage of tubule formation, 2) the degree of nuclear pleomorphism, and 3) the mitotic cell count~\cite{elston1991pathological}. 

Convolutional Neural Networks (CNNs) can be trained in an end-to-end manner to distinguish the different types of cancer, by extracting high-level information from images through stacking convolutional layers. Breast cancer classification has been fundamentally improved by the development of CNN models~\cite{wang2018classification}.
However, breast cancer segmentation in WSIs is still underexplored. WSIs are RGB images with high resolution (e.g. $80000 \times 60000$). Constrained by the memory, WSIs cannot be directly fed into the network. One solution is to crop the WSIs to small patches for patch-wise training~\cite{bandi2017comparison}. Given a fixed input size, however, there is a trade-off between accuracy and the inference speed. One can efficiently reduce the inference cost by cropping the WSIs to larger patches and rescaling the patches to a smaller input size, but this results in a loss of detail and sacrifices accuracy. In WSIs, the suspicious cancer areas our regions of interest (ROIs), are sparse, since most regions are normal tissue or the glass slide. The four classes are therefore highly imbalanced. Further, the pixel-wise annotation of breast cancer segmentation requires domain knowledge and extensive human labor and the ground truth labels are often noisy at the pixel-level. Training on patches with a small field of view can therefore easily lead to overfitting.

\begin{figure}[t]
\begin{center}
\captionsetup[subfigure]{labelformat=empty}
\subfloat[(a)]{\includegraphics[width=0.22\linewidth,height=0.22\linewidth]{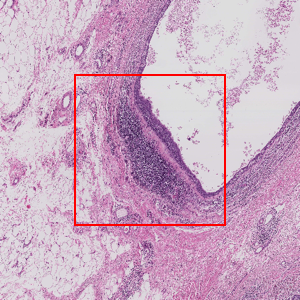}}\vspace{-0.05cm}
\subfloat[(b)]{\includegraphics[width=0.22\linewidth,height=0.22\linewidth]{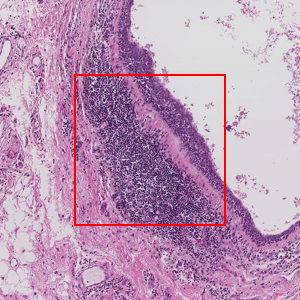}}\vspace{-0.05cm}
\subfloat[(c)]{\includegraphics[width=0.22\linewidth,height=0.22\linewidth]{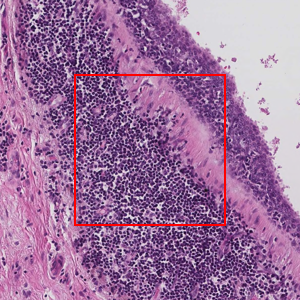}}\vspace{-0.05cm}
\subfloat[(d)]{\includegraphics[width=0.22\linewidth,height=0.22\linewidth]{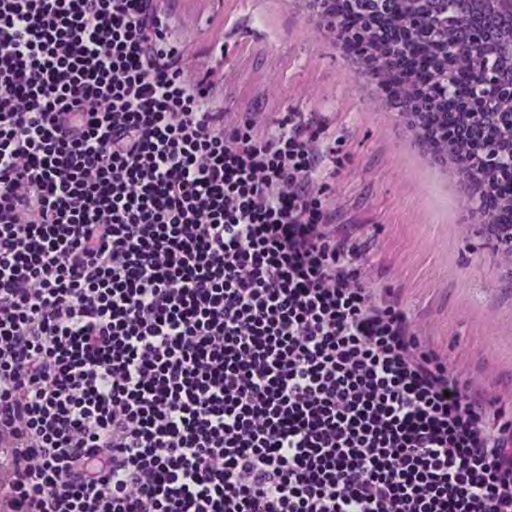}}\vspace{-0.05cm}
\end{center}
   \caption{Zoom-in process. The regions bounded by the red boxes are zoomed in sequentially with zoom-in rate 2. All zoomed-in regions are resized to the same resolution for visualization. The white regions in (a), (b) and (c) are the background glass slide.}
\label{fig:zoom}
\end{figure}

In this paper, we propose a semantic segmentation framework, Reinforced Auto-Zoom Net (RAZN). When a pathologist examines the WSIs with a digital microscope, the suspicious areas are zoomed in for details and the non-suspicious areas are browsed quickly (See Fig.~\ref{fig:zoom} for an intuition.). RAZN is motivated by this attentive zoom-in mechanism. We learn a policy network to decide the zoom-in action through the policy gradient method~\cite{sutton1998reinforcement}. By skipping the non-suspicious areas (normal tissue), noisy information (glass background) can be ignored and the WSIs can be processed more quickly. By zooming in the suspicious areas (abnormal tissue), the data imbalance is alleviated locally (in the zoomed-in regions) and more local information is considered. Combining these two can efficiently reduce overfitting for the normal tissue, which is caused by the imbalanced data, and lead to improved accuracy. However, since the zoom-in action is selective, the inference can at the same time be fast.

The previous studies on zoom-in mechanism focus on utilizing multi-scale training to improve prediction performance. The Hierarchical Auto-Zoom Net HAZN~\cite{xia2016zoom} uses sub-networks to detect human and object parts at different scales hierarchically and merges the prediction at different scales, which can be considered as a kind of ensemble learning. Zoom-in-Net~\cite{wang2017zoom} zooms in suspicious areas generated by attention maps to classify diabetic retinopathy. In both HAZN and Zoom-in-Net, the zoom-in actions are deterministic. So in the training phase, the patches will be upsampled and trained even if it may not decrease the loss. In RAZN, the zoom-in actions are stochastic, and a policy is learned to decide if the zoom-in action can improve the performance.

This paper makes the following contributions: 1) we propose an innovative framework for semantic segmentation for images with high resolution by leveraging both accuracy and speed; 2) we are the first to apply reinforcement learning to breast cancer segmentation; 3) we compare our framework empirically with multi-scale techniques used in the domain of computer vision and discuss the influence of multi-scale models for breast cancer segmentation.

\section{Reinforced Auto-Zoom Net}
In clinical practice, it is impossible for a clinician to go through each region of a WSI at the original resolution, due to the huge image size. The clinician views regions with simple patterns or high confidence quickly at coarse resolution and zooms in for the suspicious or uncertain regions to study the cells at high resolution. The proposed RANZ simulates the examining process of a clinician diagnosing breast cancer on a WSI. Another motivation of RAZN is that the characteristics of the cancer cells have different representations at different field of view. For semantic segmentation tasks on common objects, the objects in the same category share discriminative features and attributes. For example, we can differentiate a cat from a dog based on the head, without viewing the whole body. However, in cancer segmentation, the basic unit is the cell, which consists of nucleus and cytoplasm. The difference between the cells is not obvious. Instead of checking only a single cell, the diagnosis is based on the features of a group of cells, such as the density, the clustering and the interaction with the environment. RANZ is designed to learn this high-level information. 

\begin{figure}[t]
\begin{center}
\includegraphics[width=0.9\linewidth]{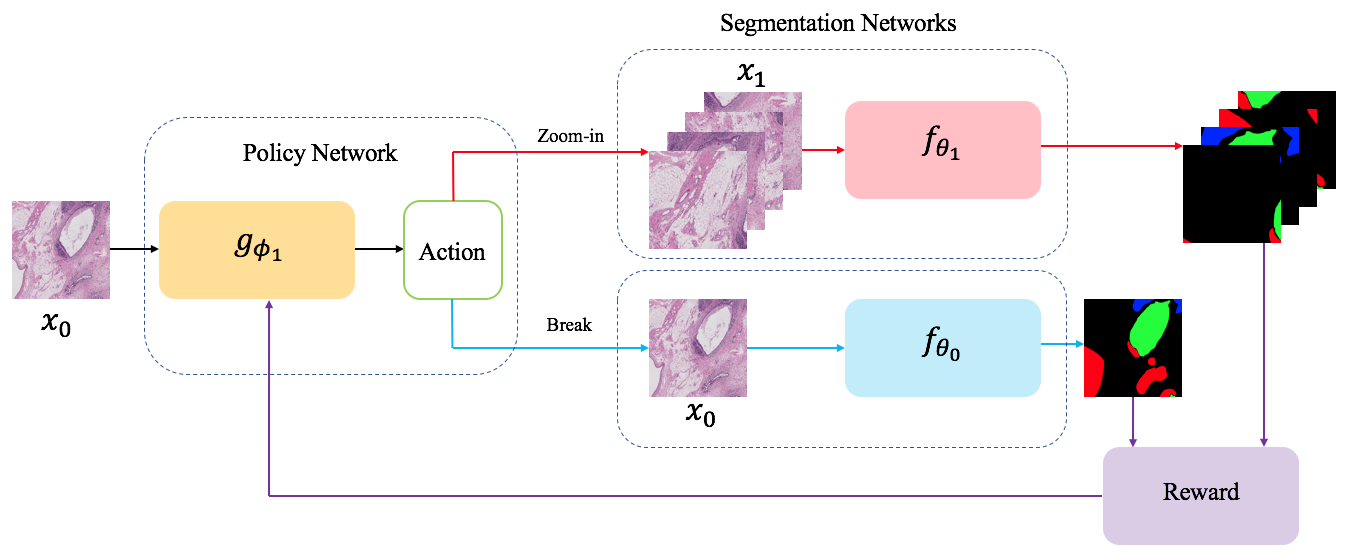}
\end{center}
   \caption{Illustration of the proposed framework when $m = 1$ and $r = 2$. In the inference phase, given a cropped image $x_0$, the policy network outputs the action, zoom-in (red arrows) or break (blue arrows). In the training phase, the policy network will be optimized to maximize the reward (purple arrows), which is determined by the segmentation prediction.}
\label{fig:arch}
\end{figure}

RAZN consists of two types of sub-networks, policy networks $\{f_\theta\}$ and segmentation networks $\{g_\phi\}$. Assume the zoom-in actions can be performed at most $m$ times and the zoom-in rate is $r$. There is one base segmentation network $f_{\theta_0}$ at the coarsest resolution. At the $i$th zoom-in level, there is one policy network $g_{\phi_i}$ and one segmentation network, $f_{\theta_i}$. In the inference time, with fixed field of view and magnification level, we have a cropped patch $x_0$ with shape $[H,W,3]$, like Fig.~\ref{fig:zoom} (a). Then $g_{\phi_1}$ will take $x_0$ as an input and predict the action, \textit{zoom-in} or \textit{break}. If the predicted action is break, $f_{\theta_0}(x_0)$ will output the segmentation results and the diagnosis for $x_0$ is finished. If the predicted action is zoom-in, a high-magnification patch $\bar{x}_0$ with corresponding zoom-in rate will be retrieved from the original image. $\bar{x}_0$, with shape $[rH,rW,3]$, will be cropped into $x_1$, which is $r^2$ patches of shape $[H,W,3]$. Then each patch of $x_1$ will be treated as $x_0$ for the next level of zoom-in action. Fig.~\ref{fig:zoom} (b) is a central crop of $x_1$. The process is repeated recursively until a pre-defined maximum magnification level is reached. In this work, we propose this novel idea and focus on the situation of $m = 1$. $m > 1$ will be discussed in future work. An overview of the architecture is illustrated in Fig.~\ref{fig:arch}.

The segmentation networks are Fully Convolutional Networks (FCNs) \cite{long2015fully} and share the same architecture. However, unlike parameter sharing in the common multi-scale training in semantic segmentation~\cite{chen2018deeplab}, each network is parameterized by independent $f_\theta$, where $f_{\theta_i}: \mathbb{R}^{H \times W \times 3} \rightarrow \mathbb{R}^{H \times W \times C}$ and $C$ is the number of classes. The reason for choosing independent networks for each zoom-in level is that CNNs are not scale-invariant \cite{goodfellow2016deep}.
Each FCN can thus learn high-level information at a specific magnification level. Given input image $x$ and segmentation annotation $y$, the training objective for each FCN is to minimize 

\begin{equation}
    J_{\theta_i}(x, y) = - \frac{1}{HW}\sum_{j}\sum_{c} y_{j,c}\log f_{\theta_i}(x)_{j,c} \; ,
\label{eq:1}
\end{equation}
where $j$ ranges over all the $H\times W$ spatial positions and $c \in \{0,...,3\}$ represents the semantic classes (cancer type).

At $m = 1$, the framework is a single-step Markov Decision Process (MDP) and the problem can be formulated by the REINFORCE rule \cite{williams1992simple}. The policy network projects an image to a single scalar, $g_{\phi_1}: \mathbb{R}^{H \times W \times 3} \rightarrow \mathbb{R}$. Given the state $x_0$, the policy network defines a policy $\pi_{\phi_1}(x_0)$. The policy samples an action $a \in \{0, 1\}$, which represents break and zoom-in, respectively. We have
\begin{equation}
    \textit{p} = \sigma(g_{\phi_1}(x_0)) \;,
\label{eq:2}
\end{equation}
\begin{equation}
    \pi_{\phi_1}(x_0) = \textit{p}^a (1-\textit{p})^{1-a} \; ,
\label{eq:3}
\end{equation}
where $\sigma(\cdot)$ is the sigmoid function and $\pi_{\phi_1}(x_0)$ is essentially a Bernoulli distribution. The motivation of RAZN is to improve the segmentation performance and it is therefore natural to define the reward such that it minimizes the segmentation loss. Based on Equation~\ref{eq:1}, we have $J_{\theta_0}(x_0, y_0)$, $J_{\theta_1}(x_1, y_1)$, where $x_1$ is the transformed $x_0$ after zoom-in and cropping operations. It is practical in reinforcement learning training to utilize the advantage function to reduce variance \cite{rennie2017self} and we therefore define the reward as 

\begin{equation}
     \textit{R}(a) = a \frac{J_{\theta_1}(x_1, y_1) - J_{\theta_0}(x_0, y_0)}{J_{\theta_0}(x_0, y_0)}.
\label{eq:4}
\end{equation}

So when $a = 1$, the reward is positive if $J_{\theta_1}(x_1, y_1) >  J_{\theta_0}(x_0, y_0)$,  and the reward is negative if $J_{\theta_1}(x_1, y_1) < J_{\theta_0}(x_0, y_0)$ . The denominator in Equation~\ref{eq:4} functions as a normalizer to prevent reward explosion. To prevent $\textit{p}$ from saturating at the beginning, we adopt the bounded Bernoulli distribution

\begin{equation}
    \tilde{\textit{p}} = \alpha \textit{p} + (1 - \alpha) (1 - \textit{p}). 
\label{eq:5}
\end{equation}

We have $\tilde{\textit{p}} \in [1-\alpha, \alpha]$. The training objective is to maximize the expected reward or to minimize the negative expected reward 

\begin{equation}
    J_{\phi_1}(x_0) = - \mathbb{E}_{a \sim \pi_{\phi_1}(x_0)}[\textit{R}(a)].
\label{eq:6}
\end{equation}

The optimization of the policy network is implemented through policy gradient methods \cite{williams1992simple,sutton1998reinforcement,sutton2000policy}, where the expected gradients are

\begin{equation}
    \frac{\partial}{\partial \phi_1} J_{\phi_1}(x_0) = - \mathbb{E}_{a \sim \pi_{\phi_1}(x_0)} [\textit{R}(a) \frac{\partial}{\partial \phi_1} \text{log}(a \tilde{\textit{p}} + (1 - a)(1 - \tilde{\textit{p}}))]
\label{eq:7}
\end{equation}
We adopt an alternating training strategy to update both networks. The training procedure of RAZN is illustrated in Algorithm~\ref{algo:1}.

\begin{algorithm}[t]
\caption{Training of RAZN when $m = 1$}
\label{algo:1}
\begin{algorithmic}[1]
\Require $x_0$ 
\State Get $J_{\theta_0}(x_0, y_0)$ and $J_{\theta_1}(x_1, y_1)$ 
\State Sample action $a$ through $\pi_{\phi_1}(x_0)$
\State Get $\textit{R}(a)(x_0)$
\State Update $\phi_1$ by minimizing $J_{\phi_1}(x_0)$
\If{$a = 1$}
\State Update $\theta_1$ by minimizing $J_{\theta_1}(x_1, y_1)$ 
\Else
\State Update $\theta_0$ by minimizing $J_{\theta_0}(x_0, y_0)$
\EndIf
\end{algorithmic}
\end{algorithm}

\section{Experiments}
\paragraph{\bf{Dataset}}
The dataset used in this study is provided by Grand Challenge on Breast Cancer Histology Images \footnote{https://iciar2018-challenge.grand-challenge.org/dataset}. The dataset contains 10 high-resolution WSIs with various image size. WSIs are scanned with Leica SCN400 at $\times 40$ magnification.
The annotation was performed by two medical experts. As annotation of WSIs requires a large amount of human labor and medical domain knowledge, only sparse region-level labels are provided and annotations contain pixel-level errors. In this dataset, the white background (glass slide) is labeled as \textit{normal} by the annotators. The dataset is unbalanced for the four cancer types. 

\paragraph{\bf{Implementation}}
Experiments are conducted on a single NVIDIA GTX Titan X GPU. In this study, $m = 1$, $r = 2$ and $\alpha = 0.8$. The backbone of $f_{\theta_i}$ is ResNet18 \cite{he2016deep}, with no downsampling performed in conv3\_1 and conv4\_1. $g_{\phi_1}$ is also based on the ResNet18 architecture. However, each block (consisting of 2 residual blocks \cite{he2016deep}) is replaced by a $3 \times 3$ convolution followed by batch normalization and ReLU non-linearity. 
The computational cost for the policy network is $7.1\%$ of the segmentation networks. The input size to the segmentation networks and the policy network is fixed to $256 \times 256$. We use the Adam optimizer \cite{kingma2015adam} for both the policy network and segmentation networks and use a step-wise learning rate policy with decay rate 0.1 every 50000 iterations. The initial learning rate is 0.01.
 
\paragraph{\bf{Multi-scale}}
Given a $256 \times 256$ patch, we consider two resolutions in order to simulate the zoom-in process. A coarse resolution (Scale 1), where the patch is downsampled to $64 \times 64$ and a fine resolution patch (Scale 2), where the patch is downsampled to $128 \times 128$. The patches are then resized back to $256 \times 256$ using bilinear interpolation. To evaluate the efficiency of the proposed framework, we compare our model with two multi-scale models. The first multi-scale model is the segmentation network $f_\theta$ with multi-scale training \cite{chen2018deeplab}, denoted as MS. We only consider two scales in this experiment (Scale 1 and Scale 2). Similarly, another multi-scale model is the multi-scale fusion with attention \cite{chen2016attention}, which is denoted as Attention. The training details of all models are the same. All models are trained with 200000 batches.

\begin{table}[t]
\centering
{\setlength{\tabcolsep}{1pt}
\begin{tabular}{ccccc|c}
\hline
 & non-carcinoma & carcinoma & mIOU & Weighted IOU & Relative Inference Time\\ \Xhline{4\arrayrulewidth}
Scale 1 & 0.45 & 0.32 & 0.38 & 0.07 & 1.00 \\ \hline
Scale 2 & 0.46 & 0.31 & 0.39 & 0.07 & 4.01 \\ \hline
MS \cite{chen2018deeplab} & 0.32 & 0.04 & 0.18 & 0.01 &5.06\\ \hline
Attention \cite{chen2016attention} & 0.43 & 0.29 & 0.36 & 0.06 & 5.16\\ \hline
RAZN & 0.49 & 0.49 & 0.49 & 0.11 & 2.71 $\pm$ 0.57 \\ \hline
\end{tabular}
}
\caption{Comparison of the performance. Non-carcinoma includes \textit{normal} and \textit{beign}. Carcinoma includes \textit{in situ carcinoma} and \textit{invasive carcinoma}.}
\label{tab:1}
\end{table}

\paragraph{\bf{Performance}}
We compare two key indicators of the performance, which are the segmentation performance and the inference speed. We use intersection over union (IOU) as the metric for segmentation performance. We report mean IOU, which is just the average IOU among four classes. Due to the imbalanced data, we also report weighted IOU, where the weight is proportional to the inverse of the frequency of the labels of each class.
Further, we report relative inference time for the proposed RAZN and the baseline methods compared to the inference time for the model that only considers Scale 1. We report the average relative inference time over 100 patches. Lower values of relative inference time represent faster inference speed. The results are presented in Table~\ref{tab:1}. Note, we report the mean and the standard deviation for RAZN, as the inference time will vary depending on whether zooming is required for a given patch or not.
It can be shown that RAZN actually performs better than the single scale and the multi-scale baselines. 
MS's performance is the worst of our benchmarks. MS exaggerates the imbalance problem by augmenting the data, which can confuse the network. We also hypothesize that the cell size is not the critical factor that influences the segmentation of cancer and that MS, therefore, aims to model unnecessary information on this task.
Similarly, attention models memorize the scale of the object by fusing the results from different scales. 
However, when the object is not well-defined at certain scales, like in our task the cancer (group of dense cells), the network may learn to fit noise. 
Our results illustrate that RAZN instead is more robust when data is noisy and imbalanced, providing an overall accuracy improvement at low inference time.

\section{Discussion and Conclusions}
We proposed RAZN, a novel deep learning framework for breast cancer segmentation in WSI, that uses reinforcement learning to selectively zoom in on regions of interest. The results show that the proposed model can achieve improved performance, while at the same time reduce inference speed compared to previous multi-scale approaches. We also discuss the use of multi-scale approaches for breast cancer segmentation. We conclude that cancer cells are different from general objects due to their relative small and fixed size. Multi-scale approaches may not work for a noisy and imbalanced data.
In future work, we aim to extend the model to study the multiple zoom-in actions situation ($m>1$) and will investigate the potential of more complex segmentation backbone models to improve overall performance.

\vspace{0.3cm}
\noindent{\bf{Acknowledgements.}} We thank ICIAR 2018 Grand Challenge on Breast Cancer Histology Images for providing the data for this study.

\bibliographystyle{splncs03}
\bibliography{dlmia}
\end{document}